\def\year{2020}\relax
\documentclass[letterpaper]{article} 
\usepackage{aaai20}  
\usepackage{times}  
\usepackage{helvet} 
\usepackage{courier}  
\usepackage[hyphens]{url}  
\usepackage{graphicx} 
\usepackage{graphicx}  
\usepackage{amssymb,amsmath,amsfonts,bm}
\usepackage{bm}
\usepackage{xspace}
\usepackage{multirow}
\usepackage[table]{xcolor}
\usepackage{array, booktabs, makecell}

\def\va{{\bm{a}}}
\def\vr{{\bm{r}}}

\def\mA{{\bm{A}}}
\def\mE{{\bm{E}}}
\def\mS{{\bm{S}}}
\def\mT{{\bm{T}}}
\def\sR{{\mathbb{R}}}
\def\sZ{{\mathbb{Z}}}
\def\gL{{\mathcal{L}}}

\newcommand{\goat}{\textsc{Goat}\xspace}
\newcommand{\goatg}{\textsc{GoatGlobal}\xspace}
\newcommand{\glb}{\textsc{Global}\xspace}
\newcommand{\vgae}{\textsc{VGAE}\xspace}

\newcommand{\deepwalk}{\textsc{DeepWalk}\xspace}
\newcommand{\ntov}{\textsc{Node2Vec}\xspace}
\newcommand{\linee}{\textsc{Line}\xspace}
\newcommand{\tadw}{\textsc{tadw}\xspace}
\newcommand{\tridnr}{\textsc{TriDnr}\xspace}
\newcommand{\cene}{\textsc{cene}\xspace}
\newcommand{\cane}{\textsc{cane}\xspace}
\newcommand{\dmte}{\textsc{dmte}\xspace}
\newcommand{\walklets}{\textsc{WalkLets}\xspace}
\newcommand{\attwalk}{\textsc{AttentiveWalk}\xspace}
\newcommand{\spliter}{\textsc{splitter}\xspace}

\title{Gossip and Attend:\\ Context-Sensitive Graph Representation Learning}
\author{Zekarias T. Kefato, Sarunas Girdzijauskas\\
KTH Royal Institute of Technology\\ 
School of Electrical Engineering and Computer Science\\
Kista, Stockholm\\
$\lbrace$zekarias,sarunasg$\rbrace$@kth.se 
}

\begin{document}

\maketitle

\begin{abstract}
Graph representation learning (GRL) is a powerful technique for learning low-dimensional vector representation of high-dimensional and often sparse graphs.
Most studies explore the structure and metadata associated with the graph using random walks and employ an unsupervised or semi-supervised learning schemes.
Learning in these methods is context-free, resulting in only a single representation per node.
Recently studies have argued on the adequacy of a single representation and proposed context-sensitive approaches, which are capable of extracting multiple node representations for different contexts. This proved to be highly effective in applications such as link prediction and ranking.

However, most of these methods rely on additional textual features that require complex and expensive RNNs or CNNs to capture high-level features or rely on a community detection algorithm to identify multiple contexts of a node.

In this study we show that in-order to extract high-quality context-sensitive node representations it is not needed to rely on supplementary node features, nor to employ computationally heavy and complex models.
We propose \goat, a context-sensitive algorithm inspired by gossip communication and a mutual attention mechanism simply over the structure of the graph.
We show the efficacy of \goat using 6 real-world datasets on link prediction and node clustering tasks and compare it against 12 popular and state-of-the-art (SOTA) baselines.
\goat consistently outperforms them and achieves up to 12\% and 19\% gain over the best performing methods on link prediction and clustering tasks, respectively.
\end{abstract}


\section{Introduction}
\label{sec:intro}
GRL is a powerful tool for learning the representation of a graph.
Such a representation gracefully lends itself to a wide variety of network analysis tasks, such as link prediction, node clustering, node classification, recommendation, etc.

Naturally, users in real world networks belong to multiple contexts at a time. For instance, on interaction networks such as YouTube users usually interact (watch, like, comment and so on) with videos from different categories or topics depending on their interest. On social networks, like Facebook users tend to befriend others from multiple aspects as a result of communication over different contexts (e.g. country, school, religion, work and so on).
This property is prevalent in many other areas, such as e-commerce, drug-target interaction networks and so on.

However, in  most GRL studies, the learning is oblivious to such contexts (context-free)~\cite{DBLP:journals/corr/PerozziAS14,Grover:2016:NSF:2939672.2939754,Wang:2016:SDN:2939672.2939753,DBLP:journals/corr/PerozziKS16,Yang:2015:NRL:2832415.2832542,Pan:2016:TDN:3060832.3060886,gat2vec}. 
This is to say that all the context information is squeezed into a single (global) latent representation.
In many cases, particularly for sparse graphs this leads to the loss of important details, and hence decreased performance in network analysis tasks.

Recently, a complementary line of research has questioned the adequacy of single representations per node and pursued a context-sensitive approach~\cite{DBLP:journals/corr/abs-1905-02138,tu-etal-2017-cane,Zhang:2018:DMT:3327757.3327858,kefato2020graph}.
This approach learns multiple representations per node to capture the different contexts that a node is part of.
That is, given an anchor node, its representation changes depending on another target (context) node it is coupled with.
A context node can be sampled from a neighborhood~\cite{tu-etal-2017-cane,Zhang:2018:DMT:3327757.3327858}, community affiliations~\cite{DBLP:journals/corr/abs-1905-02138}, random walk~\cite{DBLP:journals/corr/abs-1806-01973}, and so on.
In this study we sample from a node neighborhood (nodes connected by an edge).
That is, different neighbors of a given node potentially provide its multiple contexts.
Thus, in the learning process of our approach representation of a source node  changes depending on the target (context) node it is accompanied by.
Studies have shown that context-sensitive approaches significantly outperform previous context-free SOTA methods in link-prediction task.
A related notion~\cite{DBLP:journals/corr/abs-1802-05365,DBLP:journals/corr/abs-1810-04805} in NLP has significantly improved SOTA across several NLP tasks.

In this study we propose \goat\footnote{Source code: https://github.com/zekarias-tilahun/goat} (\textbf{Go}ssip and \textbf{At}tend), a context-sensitive graph representation learning algorithm that is inspired by \emph{gossip communication protocol} and \emph{multi-way attention mechanism}~\cite{DBLP:journals/corr/SantosTXZ16}.
In order to facilitate understanding of nodes' context, \goat allows each node to gossip with each neighbor in its surrounding (context) in parallel, like the gossip communication protocol.
To this end, a node uses its neighborhood as a message to be sent to the gossiping partners.
Then, through a mutual (multi-way) attention mechanism, nodes will be allowed to learn the context that they are part of by cross examining their neighborhood against the message they received.

\begin{figure}[t!]
    \centering
    \includegraphics[scale=0.47]{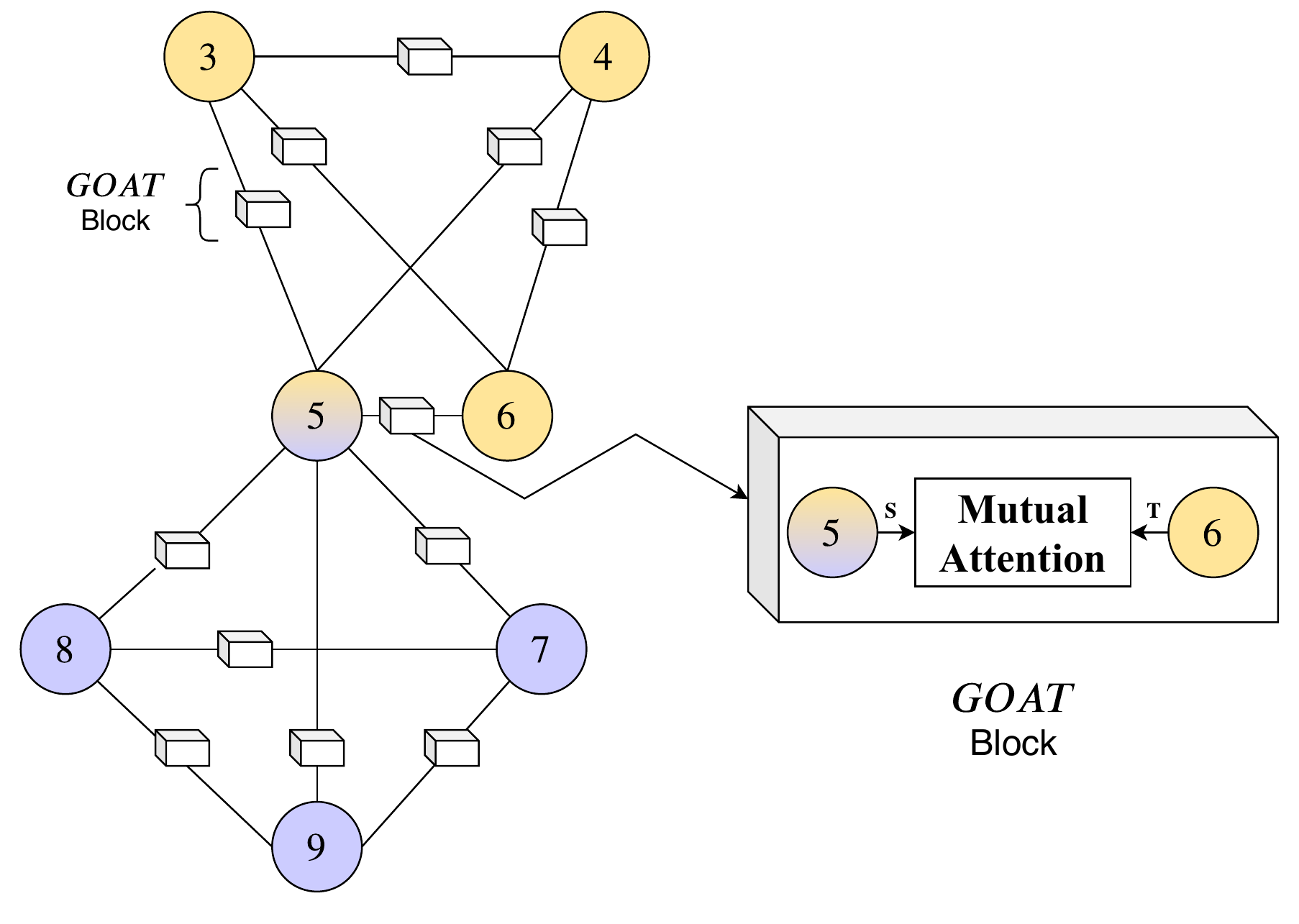}
    \caption{Left, a graph with a set of nodes having soft affiliation to groups indicated by colors and the cubes on the edges are parallel execution blocks of a shared \goat model. Right, an illustration of a \goat block for gossip partners $5$ and $6$}.
    \label{fig:goat_model}
\end{figure}

For example, in Fig.~\ref{fig:goat_model} when node 5 gossips with node 6, they both examine the message from the other node, i.e. the neighborhood set, which are ${\lbrace 3, 4, 6, 8, 9, 7 \rbrace}$ and ${\lbrace 3, 4, 5 \rbrace}$ respectively. 
In the message exchange, we want these nodes to understand that they have a shared neighborhood due to nodes \mbox{3 and 4} using the mutual-attention mechanism.
That is, we seek that both nodes, \mbox{5 and 6}, pay more attention to \mbox{3 and 4} and little attention to and/or ignore the other ones.
On the other hand, when node 5 gossips with another node, e.g. node 7, we want the attention to shift to nodes \mbox{8 and 9}.
Therefore, each time a node gossips with another context node, it pays attention to different part of its neighborhood depending on its gossip partner.
The intuition behind understanding neighborhood is reflected by changing the latent representation of a node depending on with whom it is gossiping with.
This in turn enables us to learn multiple representations per node that capture multiple facets of the node instead of just one.

Note, while \goat is inspired by gossip communication, it is not a decentralized algorithm as we monitor a central state (global embedding of nodes) and requires synchronisation.


\section{\goat}
\label{sec:goat}

\goat works over a graph ${G = (V, E)}$ with a set of $n$ nodes $V$ and $m$ edges $E$.
$G$ can be directed or undirected and weighted or unweighted. 
Without loss of generality we assume that $G$ is a weighted directed graph.
For any node ${v \in V}$, ${D(v) = \vert \lbrace u: (u, v) \in E \lor (v, u) \in E \rbrace \vert }$ denotes its degree and for every directed edge ${(s, t)}$, $w_{st}$ denotes the weight associated to the edge.

The need for the gossip-like communication between pairs of nodes stems from our objective of learning context-sensitive (multiple) representations of nodes. 
That is, by allowing nodes to independently communicate (gossip) with their neighbors we enable them to identify/understand the multiple contexts that they are part of.
For example in Fig.~\ref{fig:goat_model}, after a set of parallel gossips between 5 and other nodes in two of its contexts we want node 5 to know that it is part of two contexts as indicated by the colors of the nodes. 

However, in \goat, similar to existing studies~\cite{tu-etal-2017-cane,Zhang:2018:DMT:3327757.3327858}, the number of representations per node is the same as the number of gossip partners that it has, for example 6 representations for node 5.
This is because each gossip partner provides an understanding of a specific context that needs to be reflected by the representations.
For this reason, one has to ensure that representations of a node within a specific context are very similar to each other.
Thus, we use a mutual-attention mechanism to ensure that such representations are close to each other.
After training, one can employ nearly constant-time algorithms like \emph{locality sensitive hashing} (LSH) to collapse multiple representations of a node within the same context into a single one~\cite{Kumar_2019}.

\paragraph{Gossiping in \goat:} In the gossip-like communication that we seek to establish between a pair of nodes ${(u, v) \in E}$, the exchanged message is a neighborhood function  ${f_n: V \times \sZ \rightarrow 2^V}$, which maps each node ${u \in V}$ to a set of $N$ nodes ${N_u \subseteq V}$ sampled from the neighborhood of $u$.
A simple way of materializing $N_u$ is by sampling (without replacement) from the first-order neighbors of $u$, that is, 
\begin{equation}
    N_u = f_n(u, N) = [ v : (u, v) \in E \vee (v,u) \in E ]
\end{equation}
where $N = \vert N_u \vert $ and for the $i^{th}$ neighbor $v$, ${N_u[i] = v, \nexists j \neq i}$, where $N_u[j] = v$.
A natural assumption that we have on $N_u$ is that the order of nodes in $N_u$ has no intrinsic meaning.
In addition, though one can explore more sophisticated neighborhood sampling functions, in this study we simply consider the first order neighborhood.

Instead of the identity of its neighbors, $u_i \in N_u$, node $u$ uses a global embedding $\mE_{u_i}$ of $u_i$ to communicate with its gossip partners. 
The matrix $\mE$ defines the global (context-free) embedding of nodes and $\mE_v$ denotes the global embedding of any node $v \in V$.

Therefore given a source $s$ and a target $t$ node, where $(s, t) \in E$, the actual messages that node $s$ and $t$ use to gossip are encoded using $\mS$ and $\mT$, respectively.
\[
\mS = [\mE_{u}: u \in N_s], \quad \mT = [\mE_{v}: v \in N_t],
\]
where $\mS \in \sR^{d \times N}$, $\mT \in \sR^{d \times N}$, and $d$ is the embedding dimension. 

\paragraph{Mutual Attention} As we have alluded in an earlier discussion we use the attention mechanism so that a pair of nodes can mutually instill understanding of the shared context.

The mutual attention works over the messages from the gossiping nodes $(s, t) \in E$. Recall the messages are specified by the two $d$--by--$N$ matrices $\mS$ and $\mT$, where each column is a global embedding of the neighbors in $N_s$ and $N_t$ of $s$ and $t$, respectively.
In a nutshell, our strategy is to use an aggregated global embedding of the ``important'' neighbours to infer a context-sensitive embedding. 
Importance is quantified based on attention weights of neighbor nodes.
A neighbor's weight will be learned depending on how much information it has contained regarding the shared context of the gossipers.

\begin{figure}
    \centering
    \includegraphics[scale=0.3]{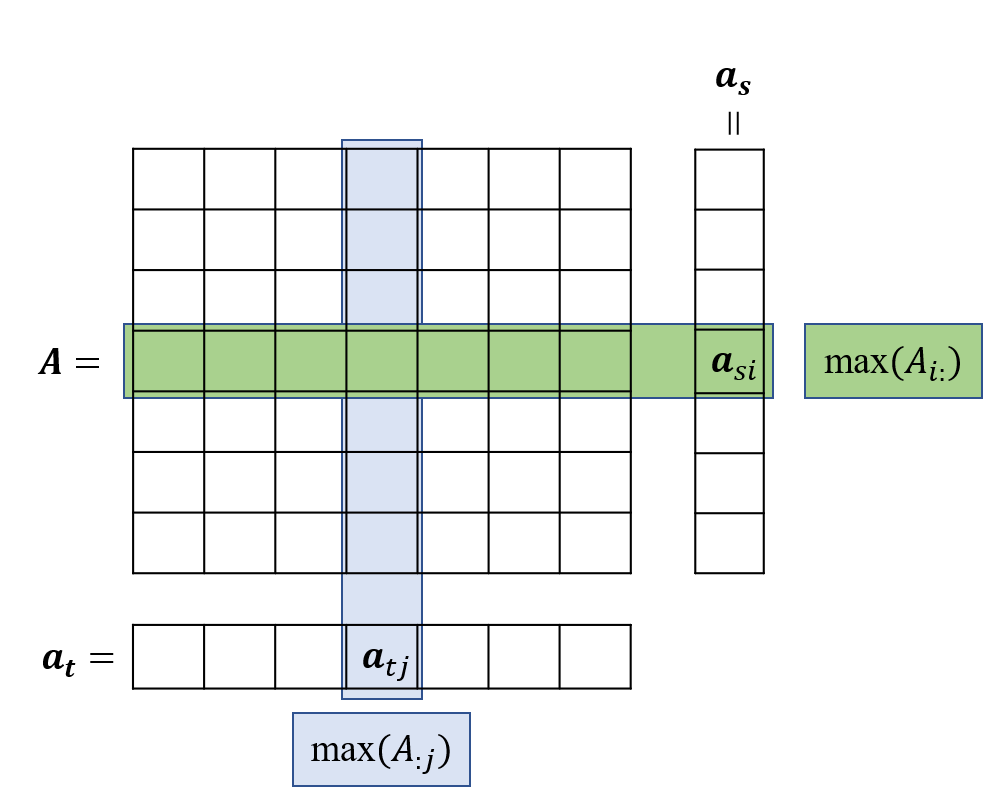}
    \caption{The alignment matrix $\mA$ and the column-wise max pooling, highlighted by the green box $\va_{si} = \max(\mA_{i:})$, and row-wise max-pooling operations, highlighted by the blue box $\va_{sj} = \max(\mA_{:j})$ to compute the unnormalized attention weight vectors $\va_s$ and $\va_t$ of nodes $s$ and $t$ respectively.}
    \label{fig:alignment_matrix}
\end{figure}
Formally, we achieve this by first computing a pair-wise soft alignment score between nodes in the set $N_s \times N_t$ as

\begin{equation}\label{eq:soft_alignment}
    \mA  = \mS^T \cdot \mT
\end{equation}

Matrix ${\mA \in \sR^{N \times N}}$ is a square matrix, where the vector in the $i^{th}$ row ${\mA_{i:} \in \sR^N}$ is associated with the $i^{th}$ neighbor ${u_i \in N_s}$ of $s$.
Each of the components $\mA_{ij}$ of $\mA_{i:}$ encode how much node $u_i$'s global embedding ${\mS^T_{u_i} = \mE_{u_i}}$ is similar/aligned to the global embeddings of each of the neighbors ${[\mE_{v_j}: v_j \in N_t]}$ of $t$.
Similarly, the vector in the $j^{th}$ column  ${\mA_{:j} \in \sR^N}$ is associated with the $j^{th}$ neighbor ${v_j \in N_t}$ of $t$. 
The components of the vector are soft alignment scores between the global embedding ${\mT_{v_j}=\mE_{v_j}}$ of $v_j$ and the global embedding of all the neighbors $[\mE_{u_i} : u_i \in N_s]$ of $s$.

Therefore, if we inspect the maximum value ${\va_{si} = \max(\mA_{i:})}$ of a particular row $\mA_{i:}$ associated with a neighbor $u_i \in N_s$ (the green box in Fig.~\ref{fig:alignment_matrix}), that will be the maximum alignment score between $u_i$ and neighbors $v_j \in N_t$ of $t$.
Thus, by examining the maximum value from all the neighbors of $s$ as $max([\va_{si} = \max(\mA_{i:}) : u_i \in N_s])$ one can tell which neighbor of $s$ is maximally aligned, \emph{i.e.}, provide information on the shared context between $s$ and $t$.
So, to identify important neighbors $N_s$ of $s$ that align with the neighbors $N_t$ of its gossip partner $t$, we perform a column-wise max-pooling operation on $\mA$ and obtain an attention weight vector $\va_s \in \sR^N$ as in Eq.~\ref{eq:column_max_pooling}. 
A similar inspection can be done on $t$ as shown by the blue box in Fig.~\ref{fig:alignment_matrix}, and ultimately attention weight vector $\va_t$ of $t$ can be computed by doing a row-wise max-pooling using Eq.~\ref{eq:row_max_pooling}.

\begin{equation}\label{eq:column_max_pooling}
    \va_s = [\max(\mA_{i:}): u_i \in N_s]
\end{equation}
\begin{equation}\label{eq:row_max_pooling}
    \va_t = [\max(\mA_{:j}): v_j \in N_t]
\end{equation}

We expect the attention weights of the important nodes, which are in the shared context of $s$ and $t$, to have higher weights and the rest to have very small weights.
Thus, if we take the weighted sum of the neighbors global-embedding, the global embedding from the important neighbors will have a stronger impact than the less important ones. 
This is exactly what we do to compute a context-sensitive representations of the gossipers.
More formally, we compute the context-sensitive representations $\vr_s \in \sR^{d \times 1}$ and $\vr_t \in \sR^{d \times 1}$ of $s$ and $t$, respectively, as the weighted sum of their neighbors global embedding using Equations~\ref{eq:context_sensitive_representation_s} and~\ref{eq:context_sensitive_representation_t}.

\begin{equation}\label{eq:context_sensitive_representation_s}
    \vr_s = \mS \cdot \texttt{softmax}(\va_s)^T
\end{equation}
\begin{equation}\label{eq:context_sensitive_representation_t}
    \vr_t = \mT \cdot \texttt{softmax}(\va_t)^T
\end{equation}\\
Softmax is used to normalize the attention weights. 
Once we devise a learning objective, the attention weights in $\va_s$ and $\va_t$ should enable us to effectively distinguish important neighbors.

\paragraph{Optimization Objective:} In order to train the aforementioned model, we employ the most commonly used learning objective in unsupervised GRL.
That is, to maximize the likelihood of the graph or the observed edges, $E$.
Concretely, we optimize the weighted negative log-likelihood of the edges specified in Eq.~\ref{eq:edge_likelihood}.

\begin{equation}\label{eq:edge_likelihood}
    \gL(E) = \min - \dfrac{1}{\vert E \vert}\sum_{(s, t) \in E} w_{st} \log P(t|s) 
\end{equation}\\
Equation~\ref{eq:edge_likelihood} seeks to minimize the negative log-likelihood of observing node $t$ as the gossip partner given node $s$, and $P(t|s)$ is estimated using the softmax formulation as follows

\begin{equation}\label{eq:soft_max}
    P(t|s) = \frac{\exp(\vr_s \cdot \vr_t)}{\sum_{w \in V}\exp(\vr_s \cdot \vr_w)}
\end{equation}\\
However, due to the normalization constant that should be computed each time a node changes a gossip partner, Eq.~\ref{eq:soft_max} is expensive to compute and we resort to negative sampling~\cite{Mikolov:2013:DRW:2999792.2999959}.
Negative nodes are sampled from the distribution $P_V$, and
a couple of alternatives, such as, the uniform and unigram distributions have been tested and~\cite{Mikolov:2013:DRW:2999792.2999959} reported that the unigram distribution raised to the power of 0.75 significantly outperforms the others, and hence we sample negative nodes according to the empirical distribution $P_V(u) = \frac{D(u)^{0.75}}{n}.$

\paragraph{Computational Complexity:} The learning in \goat is mainly affected by the number of edges, $\mathcal{O}(m)$.
The first task is to compute the embedding of each neighbor of the gossiping nodes $s$ and $t$. 
Since just a lookup is required to compute the embedding $\mE_u$ of any neighbor $u$, the cost required to embed all neighbors of the gossipers using $\mS$ and $\mT$ is $\mathcal{O}(2N)$, assuming that lookup is a constant time operation.
Second, the attention step involves a matrix multiplication given in Eq.~\ref{eq:soft_alignment}, which has a cost of $\mathcal{O}(2N^2d)$.
Therefore, the asymptotic computational cost of \goat is proportional to $\mathcal{O}(m(N^2d + 2N))$.
However, $N$ and $d$ are usually very small (less than 300) and in addition one can capitalize on highly specialized linear algebra and machine learning libraries~\footnote{We use the Numpy and PyTorch toolkits to implement \goat}.
Hence, \goat can easily scale to large graphs,
as it is mainly affected by the number of edges.
Furthermore, its design makes it easy to parallelize or decentralize the implementation.


\section{Empirical Results}
\label{sec:experiments}
\begin{table}[t!]
\centering
\begin{tabular}{@{}lccl@{}}
\hline
\textbf{Dataset} & \#\textbf{Nodes} & \#\textbf{Edges} & \textbf{Features} \\ \midrule
Cora & 2277 & 5214 & Paper Abstract \\ 
Cora2 & 2708 & 5278 & Paper Abstract \\ 
Citeseer & 3327 & 4676 & Paper Abstract \\ 
Pubmed & 19717 & 44327 & Paper Abstract \\ 
Zhihu & 10000 & 43894 & User post \\ 
Email & 1005 & 25571 & NA \\ \bottomrule
\end{tabular}
\caption{Summary of datasets, the \emph{Features} column is relevant to some of the baselines not \goat.}
\label{tbl:datasets}
\end{table}

\begin{table}[t!]
\centering
\begin{tabular}{|l|c|c|c|c|}
\hline
\textbf{Dataset} & $\rho$ & $N$ & $\lambda$ & $d$ \\ \hline
Cora (2) & \multirow{3}{*}{0.5} & \multirow{4}{*}{100} & \multirow{5}{*}{0.0001} & \multirow{5}{*}{200} \\ \cline{1-1}
Citeseer &  &  &  &  \\ \cline{1-1}
Pubmed &  &  &  &  \\ \cline{1-2}
Email & 0.8 &  &  &  \\ \cline{1-3}
Zhihu & 0.65 & 250 &  &  \\ \hline
\end{tabular}
\caption{Configuration of \goat, $\rho$ (dropout rate), $N$ (neighborhood size), $\lambda$ (learning rate), and $d$ (dimension of the latent representation).}
\label{tbl:gap_configuration}
\end{table}

In this section we provide an empirical evaluation of \goat.
To this end, we carried out experiments using the following six datasets, and a basic summary is given in Table~\ref{tbl:datasets}.
\begin{enumerate}
	\item Three of the datasets (Cora, Citeseer, and Pubmed)~\cite{tu-etal-2017-cane,Zhang:2018:DMT:3327757.3327858,kipf2016variational}: are citation network datasets, where a node represents a paper and an edge $(u, v) \in E$ represents that paper $u$ has cited paper $v$.
	For Cora we use two versions, and we refer to them as Cora and Cora2.
	\item Zhihu~\cite{tu-etal-2017-cane,Zhang:2018:DMT:3327757.3327858}: is the biggest social network for Q\&A and it is based in China. 
	Nodes are the users and the edges are follower relations between the users.
	\item Email~\cite{Leskovec:2007:GED:1217299.1217301}: is an email communication network between the largest European research institutes. A node represents a person and an edge $(u, v) \in G$ denotes that person $u$ has sent an email to $v$.
\end{enumerate}

Datasets under 1 and 2 have features (documents) associated to nodes.
Some of the baselines, discussed beneath in the 2 and 3 category, require textual information, and hence they consume the aforementioned features.
The Email dataset has ground-truth community assignment for nodes based on a person's affiliation to one of the 42 departments.

We compare our method against the following 12 popular and SOTA baselines grouped as:

\begin{enumerate}

	\item \textit{Structure based methods}:
	\deepwalk~\cite{DBLP:journals/corr/PerozziAS14}, \ntov~\cite{Grover:2016:NSF:2939672.2939754}, \walklets~\cite{DBLP:journals/corr/PerozziKS16}, \attwalk~\cite{DBLP:journals/corr/abs-1710-09599}, \linee~\cite{DBLP:journals/corr/TangQWZYM15}.
	\item \textit{Structure \& content based methods}:
	\tridnr~\cite{Pan:2016:TDN:3060832.3060886}, \tadw~\cite{Yang:2015:NRL:2832415.2832542}, \cene~\cite{DBLP:journals/corr/SunGDL16}.
	\item \textit{Structure \& content based Context-sensitive methods}: \cane~\cite{tu-etal-2017-cane}, \dmte~\cite{Zhang:2018:DMT:3327757.3327858}.
	\item \textit{Structure based context-sensitive method}: \spliter~\cite{DBLP:journals/corr/abs-1905-02138}.
	\item \textit{GCN based method}:\vgae~\cite{kipf2016variational}.
\end{enumerate}

Note that the closest algorithm to \goat is \spliter, not because of the algorithmic design but because both of them are topology (structure) based context-sensitive methods.
We also include a variant of \goat called \goatg that uses the global embedding of the nodes.
Experiments are carried out on two tasks, which are link prediction and node clustering, all of them are performed using a 24-Core CPU and 125GB RAM Ubuntu 18.04 machine.

\subsection{Link Prediction}
\label{sub:sec:link_prediction}

\begin{table*}[t!]
\centering
\begin{tabular}{|l|l|l|l|l|l|l|l|l|l|l|}
\hline
\multirow{2}{*}{\textbf{Dataset}} & \multirow{2}{*}{\textbf{Algorithm}} & \multicolumn{9}{c|}{\textbf{\% of training edges}} \\ \cline{3-11} 
 &  & 15\% & 25\% & 35\% & 45\% & 55\% & 65\% & 75\% & 85\% & 95\% \\ \hline
\multirow{14}{*}{\textbf{Cora}} & \deepwalk & 56.0 & 63.0 & 70.2 & 75.5 & 80.1 & 85.2 & 85.3 & 87.8 & 90.3 \\ \cline{3-11} 
 & \linee & 55.0 & 58.6 & 66.4 & 73.0 & 77.6 & 82.8 & 85.6 & 88.4 & 89.3 \\ \cline{3-11} 
 & \ntov & 55.9 & 62.4 & 66.1 & 75.0 & 78.7 & 81.6 & 85.9 & 87.3 & 88.2 \\ \cline{3-11} 
 & \walklets & 69.8 & 77.3 & 82.8 & 85.0 & 86.6 & 90.4 & 90.9 & 92.0 & 93.3 \\ \cline{3-11} 
 & \attwalk & 64.2 & 76.7 & 81.0 & 83.0 & 87.1 & 88.2 & 91.4 & 92.4 & 93.0 \\ \cline{2-11} 
 & \tadw & 86.6 & 88.2 & 90.2 & 90.8 & 90.0 & 93.0 & 91.0 & 93.0 & 92.7 \\ \cline{3-11} 
 & \tridnr & 85.9 & 88.6 & 90.5 & 91.2 & 91.3 & 92.4 & 93.0 & 93.6 & 93.7 \\ \cline{3-11} 
 & \cene & 72.1 & 86.5 & 84.6 & 88.1 & 89.4 & 89.2 & 93.9 & 95.0 & 95.9 \\ \cline{2-11} 
 & \cane & 86.8 & 91.5 & 92.2 & 93.9 & 94.6 & 94.9 & 95.6 & 96.6 & 97.7 \\ \cline{3-11} 
 & \dmte & 91.3 & 93.1 & 93.7 & 95.0 & 96.0 & 97.1 & 97.4 & \textbf{98.2} & \textbf{98.8} \\ \cline{2-11} 
 & \spliter & 65.4 & 69.4 & 73.7 & 77.3 & 80.1 & 81.5 & 83.9 & 85.7 & 87.2 \\ \cline{2-11} 
 & \goatg & 93.3 & 95.4 & 96.2 & 97.1 & 97.4 & \textbf{97.6} & 97.5 & 98.0 & 98.3 \\ \cline{3-11} 
 & \goat & \textbf{96.7} & \textbf{96.9} & \textbf{97.1} & \textbf{97.5} & \textbf{97.6} & \textbf{97.6} & \textbf{97.8} & 98.0 & 98.2 \\ \hline \midrule  
 \multicolumn{2}{c|}{\textbf{GAIN\%}} & 5.4\% & 3.8\% & 3.4\% & 2.5\% & 1.6\% & 0.5\% & 0.4\% &  &  \\ \midrule \hline
\multirow{14}{*}{\textbf{Zhihu}} & \deepwalk & 56.6 & 58.1 & 60.1 & 60.0 & 61.8 & 61.9 & 63.3 & 63.7 & 67.8 \\ \cline{3-11} 
 & \linee & 52.3 & 55.9 & 59.9 & 60.9 & 64.3 & 66.0 & 67.7 & 69.3 & 71.1 \\ \cline{3-11} 
 & \ntov & 54.2 & 57.1 & 57.3 & 58.3 & 58.7 & 62.5 & 66.2 & 67.6 & 68.5 \\ \cline{3-11} 
 & \walklets & 50.7 & 51.7 & 52.6 & 54.2 & 55.5 & 57.0 & 57.9 & 58.2 & 58.1 \\ \cline{3-11} 
 & \attwalk & 69.4 & 68.0 & 74.0 & 75.9 & 76.4 & 74.5 & 74.7 & 71.7 & 66.8 \\ \cline{2-11} 
 & \tadw & 52.3 & 54.2 & 55.6 & 57.3 & 60.8 & 62.4 & 65.2 & 63.8 & 69.0 \\ \cline{3-11} 
 & \tridnr & 53.8 & 55.7 & 57.9 & 59.5 & 63.0 & 64.2 & 66.0 & 67.5 & 70.3 \\ \cline{3-11} 
 & \cene & 56.2 & 57.4 & 60.3 & 63.0 & 66.3 & 66.0 & 70.2 & 69.8 & 73.8 \\ \cline{2-11} 
 & \cane & 56.8 & 59.3 & 62.9 & 64.5 & 68.9 & 70.4 & 71.4 & 73.6 & 75.4 \\ \cline{3-11} 
 & \dmte & 58.4 & 63.2 & 67.5 & 71.6 & 74.0 & 76.7 & 78.7 & 80.3 & 82.2 \\ \cline{2-11} 
 & \spliter & 59.8 & 61.5 & 61.8 & 62.1 & 62.1 & 62.4 & 61.0 & 60.7 & 58.6 \\ \cline{2-11} 
 & \goatg & 66.1 & 74.6 & 74.1 & 75.2 & 73.2 & 68.8 & 71.1 & 73.6  & 74.7 \\ \cline{3-11} 
 & \goat & \textbf{82.2} & \textbf{80.7} & \textbf{82.3} & \textbf{82.4} & \textbf{85.1} & \textbf{85.3} & \textbf{84.5} & \textbf{84.4} & \textbf{83.7} \\ \hline \midrule 
 \multicolumn{2}{c|}{\textbf{GAIN\%}} & 12.8\% & 12.7\% & 8.3\% & 6.5\% & 8.7\% & 8.9\% & 5.8\% & 4.1\% & 1.5\% \\ \midrule \hline 
\multirow{9}{*}{\textbf{Email}} & \deepwalk & 69.2 & 71.4 & 74.1 & 74.7 & 76.6 & 76.1 & 78.7 & 75.7 & 79.0 \\ \cline{3-11} 
 & \linee & 65.6 & 71.5 & 73.8 & 76.0 & 76.7 & 77.8 & 78.5 & 77.9 & 78.8 \\ \cline{3-11} 
 & \ntov & 66.4 & 68.6 & 71.2 & 71.7 & 72.7 & 74.0 & 74.5 & 74.4 & 76.1 \\ \cline{3-11} 
 & \walklets & 70.3 & 73.2 & 75.2 & 78.7 & 78.2 & 78.1 & 78.9 & 80.0 & 78.5 \\ \cline{3-11} 
 & \attwalk & 68.8 & 72.5 & 73.5 & 75.2 & 74.1 & 74.9 & 73.0 & 70.3 & 68.6 \\ \cline{2-11} 
 & \spliter & 69.2 & 70.4 & 69.1 & 69.2 & 70.6 & 72.8 & 73.3 & 74.8 & 75.2 \\ \cline{2-11} 
 & \goatg & 78.6 & 80.3 & 80.8 & 81.1 & 81.3 & 81.8 & 82.0 & 82.1 & 82.6 \\ \cline{3-11} 
 & \goat & \textbf{78.9} & \textbf{81.0} & \textbf{81.2} & \textbf{81.4} & \textbf{81.7} & \textbf{82.4} & \textbf{82.3} & \textbf{82.6} & \textbf{83.1} \\ \hline \midrule 
 \multicolumn{2}{c|}{\textbf{GAIN\%}} & 8.3\% & 5.6\% & 3.4\% & 0.8\% & 1.8\% & 2.7\% & 2.4\% & 1.5\% & 1.7\% \\ \midrule 
\end{tabular}
\caption{AUC results for the link prediction task on the Cora, Zhihu, and Email datasets.}
\label{tbl:auc_results_big_table}
\end{table*}

\begin{figure*}
    \centering
    \includegraphics[scale=0.5]{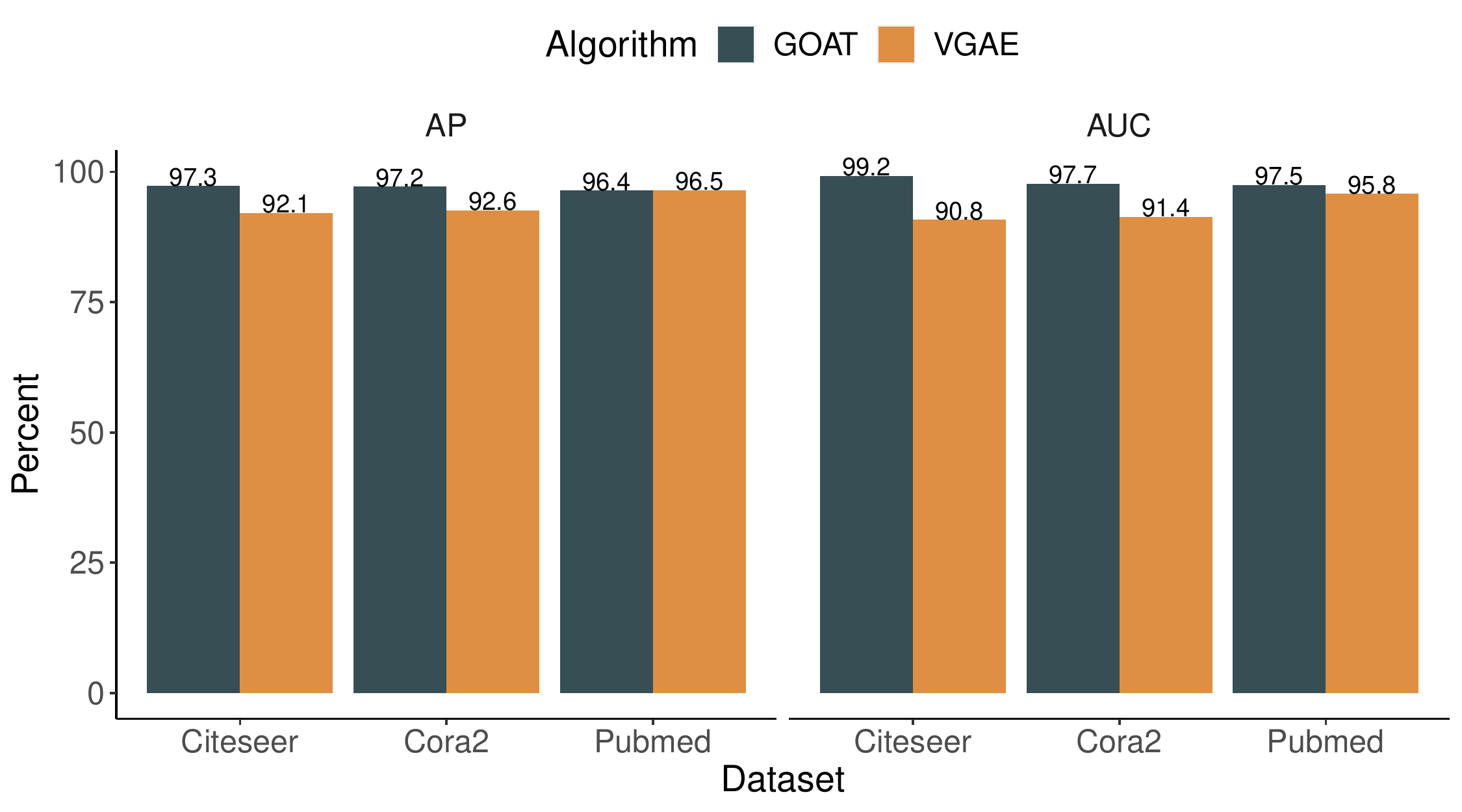}
    \caption{AUC and AP results of \vgae and \goat  for link prediction task.}
    \label{fig:vgae_vs_goat}
\end{figure*}

Link prediction is an important task that graph embedding algorithms are applied to.
Particularly context-sensitive embedding techniques have proved to be well suited for this task.
Similar to existing studies we perform this experiment using a fraction of the edges as a training set. 
We hold out the remaining fraction of the edges from the training phase and we will only reveal them during the test phase, results are reported using this set.
All hyper-parameter tuning is performed by taking a small fraction (20\%) of the training set as a validation set.

\paragraph{Setup:}
In-line with existing techniques~\cite{tu-etal-2017-cane,Zhang:2018:DMT:3327757.3327858}, the percentage of training edges ranges from 15\% to 95\% by a step of 10. 
The hyper-parameters of all algorithms are tuned using random-search.
For some of the baselines, our results are consistent with what is reported in previous studies, and hence for Cora and Zhihu we simply report these results.

Except the ``unavoidable'' hyper-parameters (eg. learning rate, regularization/dropout rate) that are common in all the algorithms, our model has just one hyper-parameter, which is the neighborhood size -- $N$, and for nodes with smaller neighborhood size we use zero padding.
As we shall verify later, \goat is not significantly affected by the choice of this parameter.

The quality of the link prediction is measured using the area under the receiver operating characteristic curve (AUC) and average precision (AP) scores.
AUC indicates the probability that a randomly selected pair $(u, w) \notin E$ will be ranked lower than an edge $(u, v) \in E$ in the test set.
The AP indicates the quality of the overall ranking as a summary of precision and recall scores at different thresholds.
Rank of a pair of nodes is computed as the dot product of their representation.
For all the algorithms the representation size -- $d$ is 200 and \goat's configuration is shown in Table~\ref{tbl:gap_configuration}.

\paragraph{Results:}
The results for Cora, Zhihu, and Email datasets are reported in Table~\ref{tbl:auc_results_big_table}.
\goat outperforms the SOTA baselines in all cases for Zhihu and Email, and in almost all cases for Cora.
One can see that as we increase the percentage of training edges, performance significantly increases for all the baselines. 
As indicated by the ``Gain'' row, \goat achieves up to 12.8\% improvement over SOTA and context-sensitive techniques.
Notably the gain is pronounced for smaller values of percentage of edges used for training.
This shows that \goat is suitable both in cases where there are several missing links and most of the links are present. 

Recently, several studies have pursued a type of GRL models known as graph convolutional neural networks (GCNs)~\cite{kipf2017semi,Velickovic2017GraphAN,DBLP:journals/corr/abs-1905-00067,DBLP:journals/corr/HamiltonYL17,DBLP:journals/corr/abs-1902-07153,kipf2016variational}.
Though most of them are widely used for semi-supervised node classification, in this study we compare \goat with a type of GCN, called variational graph auto-encoder (\vgae) that is commonly used for the link-prediction task~\cite{kipf2016variational,schlichtkrull2017modeling} using three of the remaining datasets from the \vgae paper.
For a fair comparison, we use exactly the same configuration, that is, the same training (90\%) and test (10\%) sets provided by the authors.

In Fig.~\ref{fig:vgae_vs_goat} we report the AUC and AP empirical results on Citeseer, Cora2 and Pubmed datasets, yet again we show that \goat outperforms \vgae in almost all the cases, by upto 8\%.
A consistent observation that we have in the above results is that, compared to all the baselines \goat's performance is robust even when we have little observation.

\subsection{Node Clustering}
\label{sub:sec:node_clustering}

\begin{table*}[t!]
\centering
\begin{tabular}{|l|l|l|l|l|l|l|l|l|}
\hline
\multirow{3}{*}{\textbf{Algorithm}} & \multicolumn{8}{c|}{\textbf{\%of training edges}} \\ \cline{2-9} 
 & \multicolumn{2}{c|}{\textbf{35\%}} & \multicolumn{2}{c|}{\textbf{55\%}} & \multicolumn{2}{c|}{\textbf{75\%}} & \multicolumn{2}{c|}{\textbf{95\%}} \\ \cline{2-9} 
 & \textbf{NMI} & \textbf{AMI} & \textbf{NMI} & \textbf{AMI} & \textbf{NMI} & \textbf{AMI} & \textbf{NMI} & \textbf{AMI} \\ \hline
\deepwalk & 41.3 & 28.6 & 53.6 & 44.8 & 50.6 & 42.4 & 57.6 & 49.9 \\ \cline{2-9}
\linee & 44.0 & 30.3 & 49.9 & 38.2 & 53.3 & 42.6 & 56.3 & 46.5 \\ \cline{2-9}
\ntov & 46.6 & 35.3 & 45.9 & 35.3 & 47.8 & 38.5 & 53.8 & 45.5 \\ \cline{2-9}
\walklets & 47.5 & 39.9 & 55.3 & 47.4 & 54.0 & 45.4 & 50.1 & 41.6 \\ \cline{2-9}
\attwalk & 42.9 & 30.0 & 45.7 & 36.5  & 44.3 & 35.7 & 47.4 & 38.5 \\ \hline
\spliter & 38.9 & 23.8 & 43.2 &  30.3 & 45.2 & 33.6 & 48.4 & 37.6 \\ \hline
\goat & \textbf{66.5} & \textbf{57.2} & \textbf{65.6} & \textbf{56.7} & \textbf{66.4} & \textbf{57.9} & \textbf{65.5} & \textbf{57.0} \\ \hline \hline
\textbf{\%Gain} &  \multicolumn{2}{c|}{19\%}  &  \multicolumn{2}{c|}{10.3\%} &  \multicolumn{2}{c|}{12.4\%}  &   \multicolumn{2}{c|}{7.9\%}   \\ \hline
\end{tabular}
\caption{NMI and AMI scores for node clustering experiment on the Email dataset. The Gain is with respect to the NMI only.}
\label{tbl:node_clustering_email}
\end{table*}

Nodes in a network has the tendency to form cohesive structures based on shared latent properties.
These structures are commonly known as groups, clusters or communities and their identification has important real-world applications.
We use the Email dataset since it has 42 ground truth communities.
Recall that this dataset has only structural information, thus we have included structure-based methods only.

\paragraph{Setup:} 
Since each node belongs to exactly one cluster, we employ the k-Means algorithm to identify clusters.
The learned representations of nodes by a certain algorithm are the input features of the clustering algorithm.
In this experiment the percentage of training edges varies from 35\% to 95\% by a step of 20\%, for the rest we use the same configuration as in the above experiment.

Given the ground truth community assignment $y$ of nodes and the predicted community assignments $\hat{y}$, usually the agreement between $y$ and $\hat{y}$ are measured using mutual information $I(y, \hat{y})$.
However, $I$ is not bounded and difficult for comparing methods, hence we use two other variants of $I$~\cite{Vinh2010InformationTM}.
Which are, the normalized mutual information $NMI(y, \hat{y})$, which simply normalizes $I$ and adjusted mutual information $AMI(y, \hat{y})$, which adjusts or normalizes $I$ to random chances.

\paragraph{Results:}
The results of this experiment are reported in Table~\ref{tbl:node_clustering_email}, and \goat significantly outperforms all the baselines by up to 19\% with respect to AMI score.
Consistent to our previous experiment \goat's performance is not affected by the change in the percentage of the training edges for both NMI and AMI.

\subsection{Ablation Study}
\label{sub:sec:ablation}

\begin{figure}[t!]
    \centering
    \includegraphics[scale=.55]{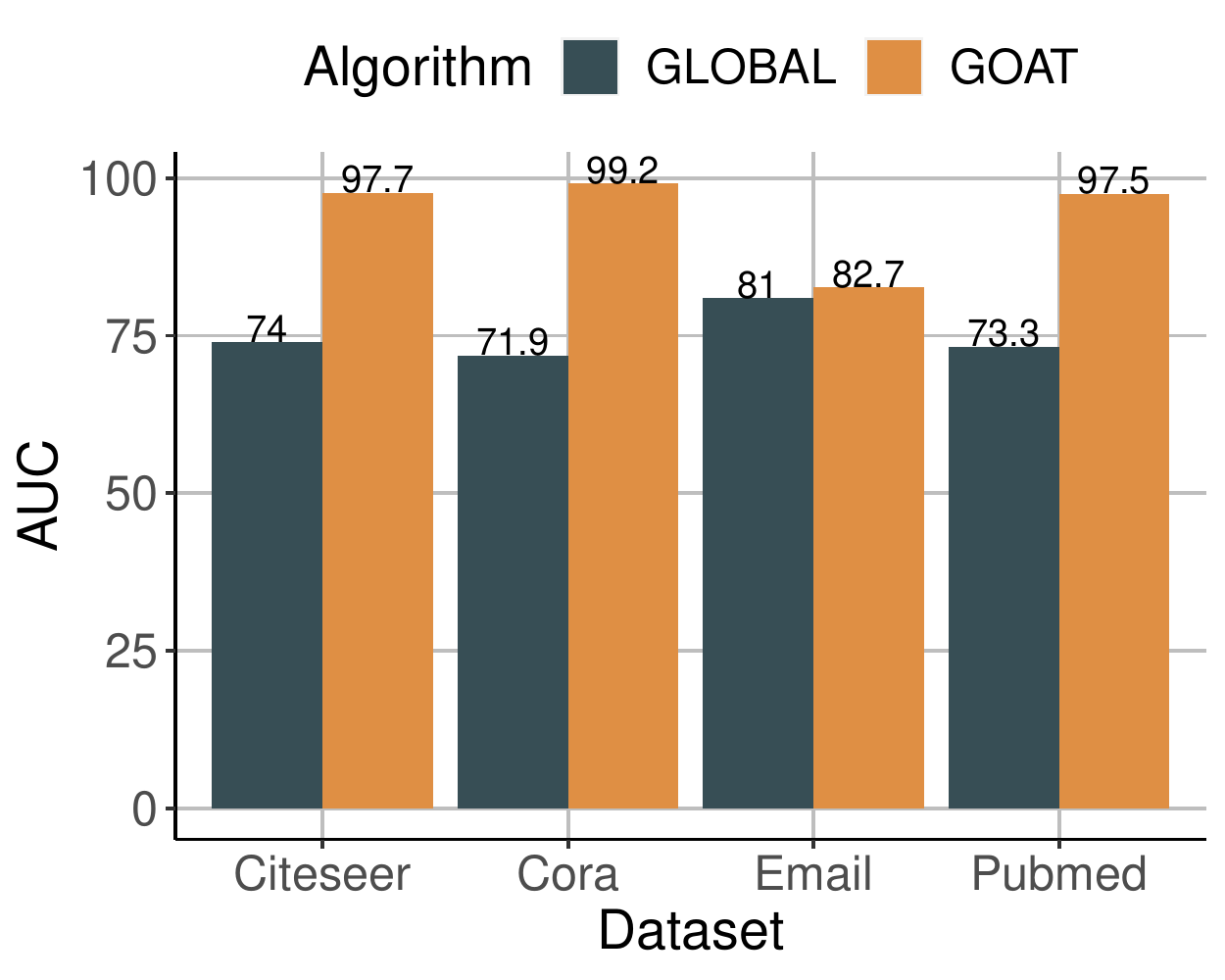}
    \caption{Comparison between \goat and \glb.}
    \label{fig:ablation}
\end{figure}

\begin{figure*}[t!]
\centering
\includegraphics[scale=0.7]{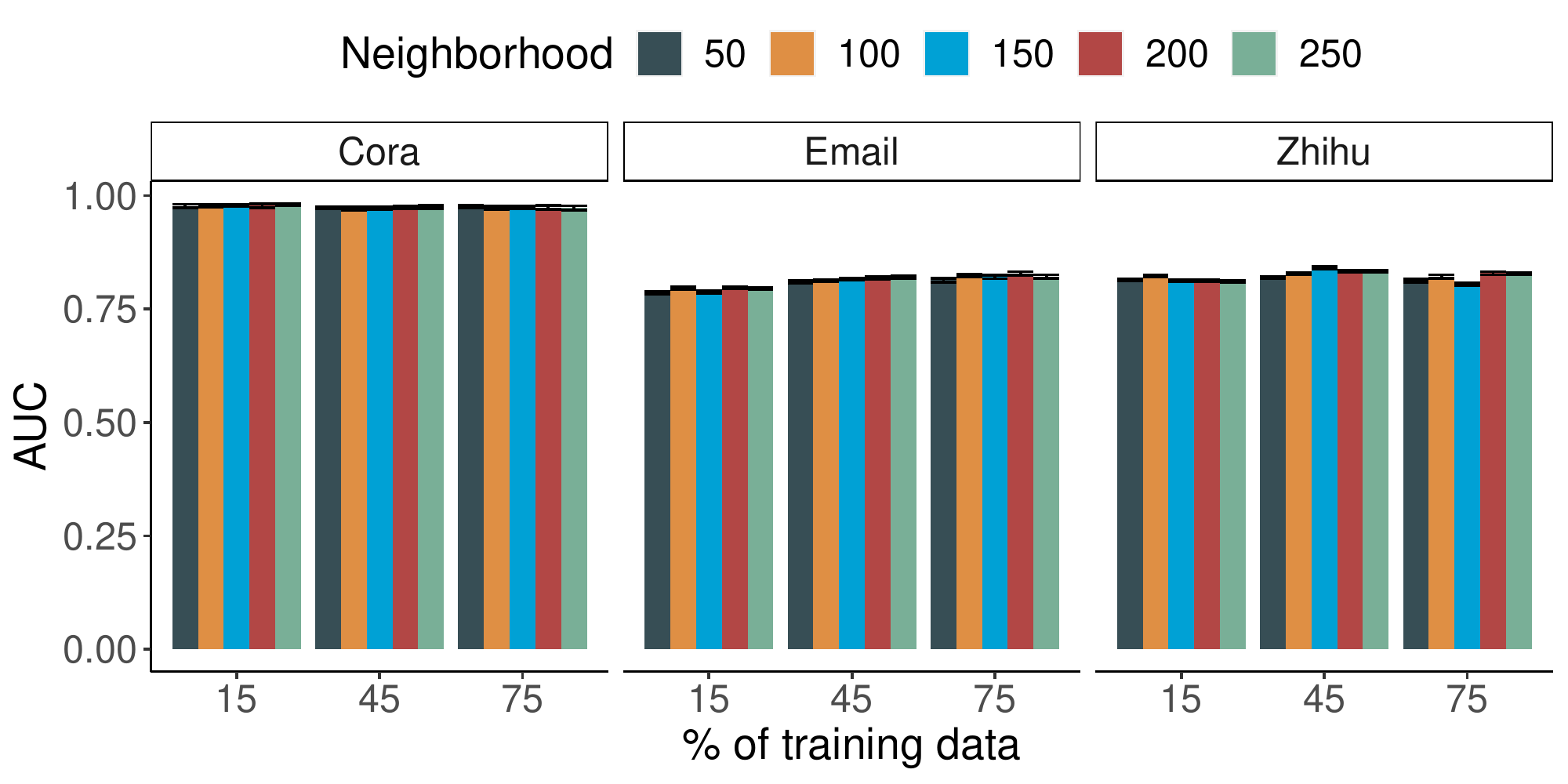}
\caption{Sensitivity of \goat to the size of node's neighborhood ($N$) on the link prediction task.}
\label{fig:neighborhood_size_sensitivity_lp}
\end{figure*}

To appreciate the importance of the mutual-attention component of \goat, we carry out an experiment by removing the attention component.
That is, instead of the context-sensitive representations $\vr_s$ and $\vr_t$, in Eq.~\ref{eq:soft_max} we use the global embeddings $\mE_s$ and $\mE_t$, and refer to this variant simply as \glb. 
It is similar to the second-order preserving variant of \linee~\cite{DBLP:journals/corr/TangQWZYM15}.
From Fig.~\ref{fig:ablation}, one can clearly see that the mutual-attention component of \goat is crucial for its effectiveness.

\subsection{Parameter Sensitivity Analysis}
\label{sub:sec:parameter_sensitivity_analysis}

Now, we turn into analyzing the effect of the main hyper-parameter of \goat, which is the size of the neighborhood ($N$).
In Fig~\ref{fig:neighborhood_size_sensitivity_lp} we show the effect of this parameter across different rate of training edges on the link prediction task.
We observe that, regardless of the percentage of training edges, \goat is not significantly affected by the change in $N$.

\subsection{Scalability and Convergence}
\begin{figure}
    \centering
    \includegraphics[scale=0.25]{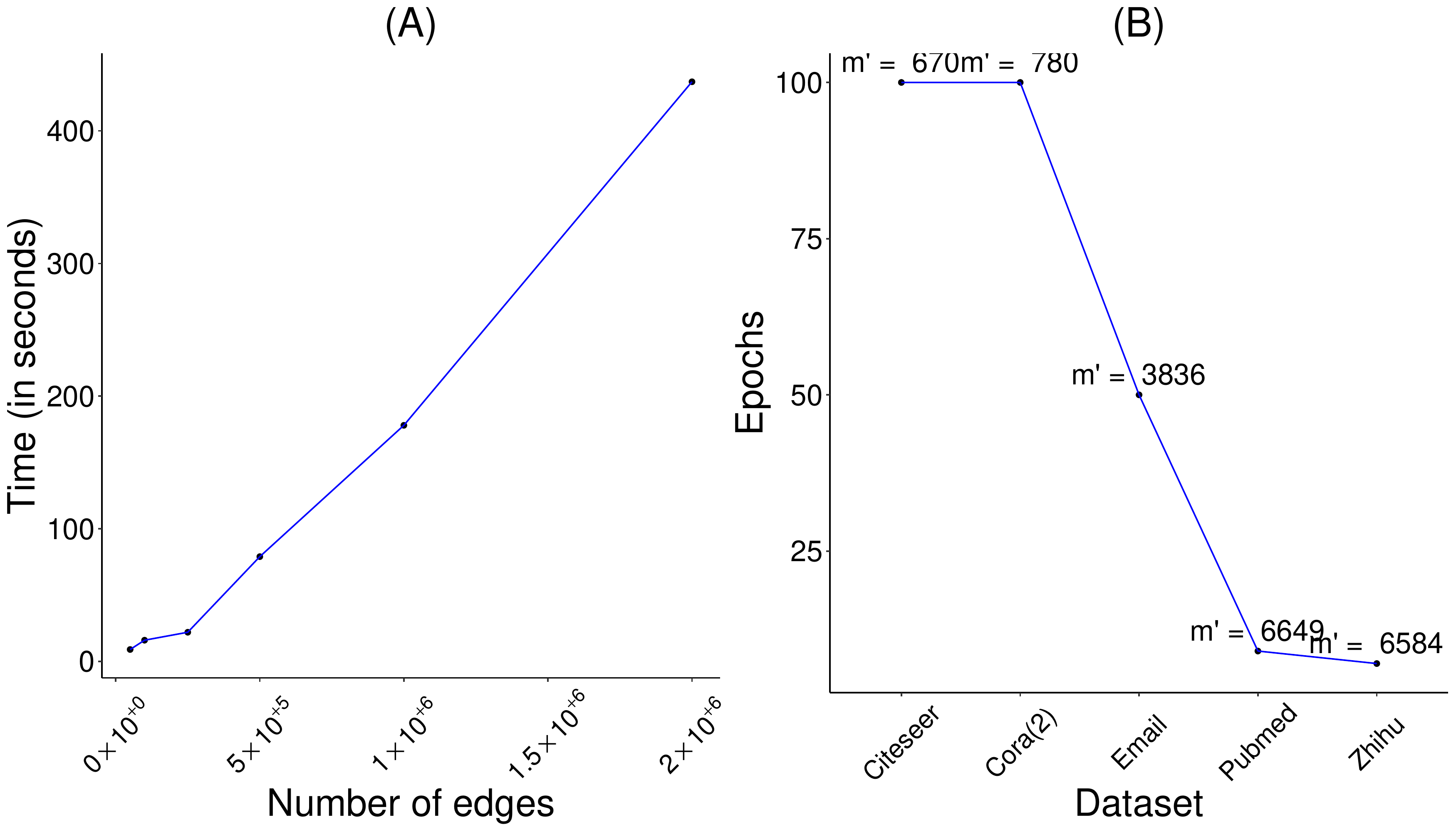}
    \caption{Run time (A) of \goat on an NVIDIA QUADRO RTX 5000 GPU and its convergence (B). The annotation in (B) shows the fraction of training edges $m' = \vert E \vert \times .15$.}
    \label{fig:goat_run_time_and_convergence}
\end{figure}

To empirically substantiate \goat's scalability, we carry out experiments on synthetic graphs up to millions of edges, which are generated using Barab\'{a}si–Albert model. Fig.~\ref{fig:goat_run_time_and_convergence}(A) shows the run time ($y$--axis) needed to complete an epoch for graphs with different number of edges, 50K-2M ($x$ -- axis), and we note that \goat can finish an epoch in $\approx$ 7 min for the graph with 2M edges. Moreover, once the model hyperparameters are fixed, we have empirically observed that for large and dense graphs \goat requires small number of epochs to converge.
Fig~\ref{fig:goat_run_time_and_convergence}(B) shows this observation, and the $y$--axis indicates the number of epochs required for \goat to converge on 15\% training edges in-order to achieve the performance reported in Section~\ref{sub:sec:link_prediction}.

\subsection{Case Study: Les Misérables}
\label{sub:sec:case_study}

\begin{figure}[t!]
    \centering
    \includegraphics[scale=0.3]{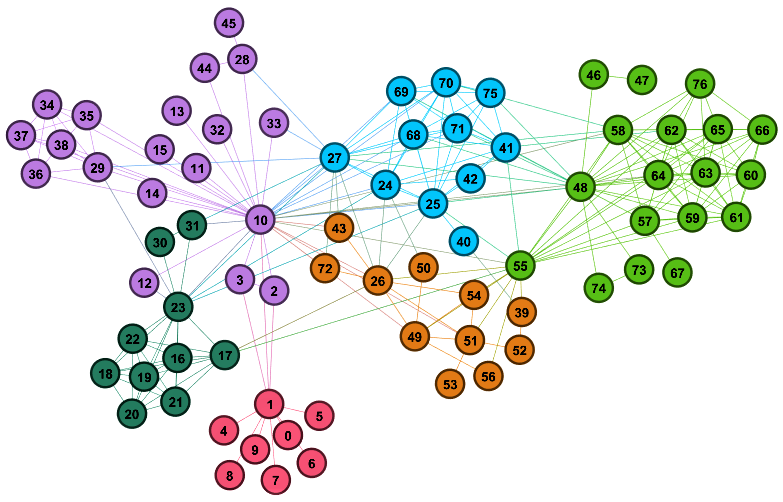}
    \caption{The Les Misérables character encounter graph with character's modularity class indicated by colors.}
    \label{fig:les_miserables}
\end{figure}
\begin{figure}[b!]
    \centering
    \includegraphics[width=.48\textwidth]{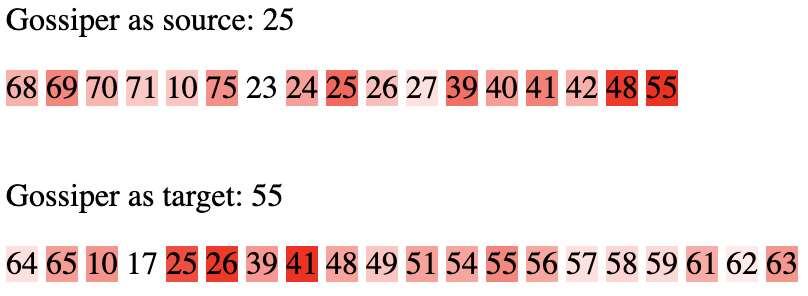}
    \caption{Visualization of the learned attention weights of neighbors of 25 and 55.}
    \label{fig:attention_vis_25_vs_55}
\end{figure}
\begin{figure}[b!]
    \centering
    \includegraphics[width=.48\textwidth]{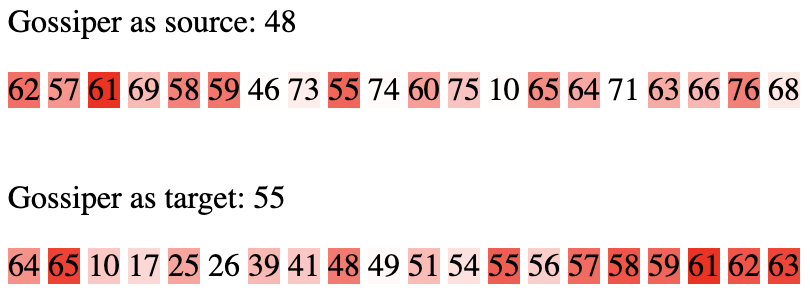}
    \caption{Visualization of the learned attention weights of neighbors of 48 and 55.}
    \label{fig:attention_vis_48_vs_55}
\end{figure}

To shed more light on the \goat model, here we briefly analyze the learned attention weights of a graph based on the encounter relations between the characters in the Les Misérables novel~\cite{Knuth:1993:SGP:164984}.
A visualization is shown in Fig~\ref{fig:les_miserables}, along with community affiliations (contexts) of nodes based on modularity classes, indicated by the colors.
First, we pick two arbitrary nodes, which are nodes $25$ and $55$.
In Fig~\ref{fig:attention_vis_25_vs_55}, we observe that both 25 and 55 strongly pay attention to the shared neighbors.
Then, we let 55 to gossip with another node, \emph{i.e.} 48, in a different context and show the resulting attention weights in Fig~\ref{fig:attention_vis_48_vs_55}.
Now, for neighbors of 55 we observe that the attention weight is concentrated on those neighbors that had less attention in the previous gossip.

\section{Related Work}
\label{sec:related_work}
Graph Representation Learning is usually carried out by exploring the structure of the graph and meta data, such as node attributes, attached to the graph~\cite{DBLP:journals/corr/PerozziAS14,Grover:2016:NSF:2939672.2939754,DBLP:journals/corr/TangQWZYM15,DBLP:journals/corr/PerozziKS16,Wang:2016:SDN:2939672.2939753,Yang:2015:NRL:2832415.2832542,Pan:2016:TDN:3060832.3060886,gat2vec}. 
Random walks are widely used to explore local/global neighborhood structures, which are then fed into a learning algorithm.
Often, an unsupervised learning objective is specified using the maximum likelihood of neighboring nodes/attributes given a center node.

Recently, graph convolutional networks have also been proposed for semi-supervised network analysis tasks~\cite{kipf2017semi,DBLP:journals/corr/HamiltonYL17,DBLP:journals/corr/abs-1902-07153,Velickovic2017GraphAN,DBLP:journals/corr/abs-1905-00067}.
These algorithms are trained to learn different kinds of neighborhood feature aggregator functions using a down-stream objective based on partial labels of nodes.
All these methods are essentially different from our approach because they are context-free.

Context-sensitive learning is another paradigm for NRL that challenges the adequacy of a single representation of a node for applications such as, link prediction, product recommendation, ranking.
While some of these methods~\cite{tu-etal-2017-cane,Zhang:2018:DMT:3327757.3327858} rely on textual information, others have also shown that a similar goal can be achieved using just the structure of the graph~\cite{DBLP:journals/corr/abs-1905-02138}. 
However, they require an extra step of persona decomposition that is based on microscopic level community detection algorithms to identify multiple contexts of a node.
Besides, it is susceptible to errors propagating from wrong community assignments.
Unlike the former approaches our algorithm does not require extra textual information and with respect to the later our approach does not require any sort of community detection algorithm.

\section{Conclusion}
\label{sec:conclusion}
In this study, we present a novel context-sensitive graph embedding algorithm called \goat.
\goat is inspired by a gossip-like communication and mutual attention mechanism.
Each node is allowed to gossip with each neighbor by using the remainder of the neighborhood as a message.
By capitalizing on the mutual attention mechanism \goat allows nodes to understand their contexts and infer multiple representations per node.

\goat learns high-quality context-sensitive representations of nodes. 
We have empirically evaluated the quality of the representations and have shown that it consistently outperforms best performing SOTA context-sensitive and context-free baselines using 6 public datasets in link prediction and node clustering tasks, exhibiting significant improvements of up to 12\% and 19\% respectively.

In a future work we seek to extend GOAT to a completely decentralized environment and investigate how node attributes can be integrated in the \goat framework.

\bibliography{aaai}
\bibliographystyle{aaai}

\end{document}